\documentclass[9pt,conference]{IEEEtran}
\IEEEoverridecommandlockouts

\usepackage{amsmath,amsfonts,bm}

\def\eqref#1{equation~\ref{#1}}

\def\1{\bm{1}}

\DeclareMathAlphabet{\mathsfit}{\encodingdefault}{\sfdefault}{m}{sl}
\SetMathAlphabet{\mathsfit}{bold}{\encodingdefault}{\sfdefault}{bx}{n}

\usepackage[utf8]{inputenc}
\usepackage[T1]{fontenc}
\usepackage{url}
\usepackage{booktabs}    
\usepackage{amsfonts}     
\usepackage{nicefrac}       
\usepackage{microtype}      
\usepackage{xcolor}       
\usepackage[pagebackref,breaklinks,colorlinks,citecolor=blue]{hyperref}
\usepackage[ruled,vlined,linesnumbered]{algorithm2e}
\usepackage{amsmath}
\usepackage{multirow}
\usepackage{multicol}

\usepackage{amssymb}
\usepackage{longtable}
\usepackage{graphicx}
\usepackage{natbib}
\usepackage{csvsimple}
\usepackage{rotating}
\usepackage{diagbox}
\usepackage{tabularx}

\usepackage[most]{tcolorbox}
\setcitestyle{numbers,square,comma}

\newcommand{\red}[1]{\textcolor{red}{#1}}

\title{Boosting Jailbreak Attack with Momentum}

\author{\IEEEauthorblockN{Yihao Zhang${}^*$}
\IEEEauthorblockA{\textit{Peking University} \\
Beijing, China \\
\texttt{zhangyihao@stu.pku.edu.cn}
\thanks{${}^*$Equal contribution.}
}
\and
\IEEEauthorblockN{Zeming Wei${}^{*\dagger\ddagger}$}
\IEEEauthorblockA{\textit{Peking University} \\
Beijing, China \\
\texttt{weizeming@stu.pku.edu.cn}
\thanks{${}^\dagger$Corresponding author.}
\thanks{${}^\ddagger$Work done during visiting UC Berkeley.}
}
}

\begin{document}

\maketitle

\begin{abstract}
Large Language Models (LLMs) have achieved remarkable success across diverse tasks, yet they remain vulnerable to adversarial attacks, notably the well-known jailbreak attack. In particular, the Greedy Coordinate Gradient (GCG) attack has demonstrated efficacy in exploiting this vulnerability by optimizing adversarial prompts through a combination of gradient heuristics and greedy search. However, the efficiency of this attack has become a bottleneck in the attacking process. To mitigate this limitation, in this paper we rethink the generation of the adversarial prompts through an optimization lens, aiming to stabilize the optimization process and harness more heuristic insights from previous optimization iterations. Specifically, we propose the \textbf{M}omentum \textbf{A}ccelerated G\textbf{C}G (\textbf{MAC}) attack, which integrates a momentum term into the gradient heuristic to boost and stabilize the random search for tokens in adversarial prompts. Experimental results showcase the notable enhancement achieved by MAC over baselines in terms of attack success rate and optimization efficiency. Moreover, we demonstrate that MAC can still exhibit superior performance for transfer attacks and models under defense mechanisms. Our code is available at \url{https://github.com/weizeming/momentum-attack-llm}.
\end{abstract}

\section{Introduction}

The discovery of adversarial examples~\citep{szegedy2013intriguing,goodfellow2014explaining} for modern deep learning models has raised continuous concern for their deployment, yet the defense for these adversarial attacks remains an open research problem~\citep{carlini2017adversarial, athalye2018obfuscated,croce2020reliable,chen2023robust,chen2024your}. Moreover, with the milestone success of Large Language Models (LLMs), they have also received serious safety concerns for this vulnerability against malicious usage, which is typically referred to as the \textit{jailbreak} attack~\citep{wei2023jailbroken,shen2023do,dong2023robust}. To tackle this issue, numerous efforts have been dedicated to fine-tuning the pre-trained language models to reduce their generation toxicity, which is considered a part of the alignment process~\citep{ouyang2022training,bai2022constitutional}.

Nevertheless, as demonstrated by Zou et al. \cite{zou2023universal}, LLMs remain susceptible to gradient-based attacks, wherein adversaries manipulate prompts to induce the LLM to generate harmful or undesirable content. Drawing inspiration from \textbf{AutoPrompt} \citep{shin2020autoprompt}, which employs gradient heuristics and search techniques to automatically generate prompts for eliciting knowledge from LLMs, the proposed Greedy Coordinate Gradient (\textbf{GCG}) attack \citep{zou2023universal} optimizes an adversarial suffix for a given malicious prompt (\textit{e.g.}, \texttt{how to build a bomb}) using gradient information and greedy search. By attaching the suffix to the target harmful request, GCG can effectively circumvent various popular LLMs. Furthermore, one intriguing property of the GCG attack is the \textit{universality} that the crafted adversarial suffix can work across different prompts. Specifically, by crafting an adversarial suffix $s$ from optimizing over a batch of malicious prompts $\{p_1,\cdots,p_n\}$ (referred to as \textit{training set}), such suffix $s$ can also be used to jailbreak unseen prompts.
Besides, several concurrent studies \citep{yong2024lowresource,yuan2024gpt} have introduced black-box attack methods that do not rely on access to input gradients or model parameters, underscoring the urgency of addressing this safety concern.

However, such a gradient-based attack encounters efficiency bottlenecks, with the optimization process being notably time-consuming. The default optimization epochs for the GCG attack typically span 500 steps, each involving numerous forward passes, resulting in tens of minutes required to attack a single prompt. To address this challenge, in this paper we rethink the attack process from an optimization standpoint. Specifically, each iteration of the GCG attack can be conceptualized as a step of gradient descent over the heuristic loss. Drawing inspiration from stochastic gradient descent (SGD) and momentum methods, we demonstrate that the optimization of the suffix can be expedited by incorporating a momentum term, akin to widely employed techniques in modern deep learning. Based on these insights, we introduce our novel gradient-based attack as the \textbf{{M}omentum {A}ccelerated G{C}G ({MAC})} attack, distinguished by the inclusion of a momentum term to enhance the optimization process. The MAC attack dynamically adjusts the adversarial suffix following each forward-backward pass, ensuring stability across diverse prompts in the training set. 

We further conduct experiments for both the individual and multiple prompt attack cases to showcase the notable acceleration achieved by MAC, along with an improvement in the attack success rate. For example, MAC can achieve a higher multiple attack success rate (ASR) of 48.6\% on vicuna-7b~\cite{zheng2023judging} with only 20 steps, significantly higher than vanilla GCG (38.1\%). Additionally, we perform in-depth investigations into MAC regarding transfer attacks. As adversarial suffices generated by GCG can transfer between models to execute black-box attacks using surrogate models, we demonstrate that MAC exhibits better transferability facilitated by the momentum term. Lastly, since some preliminary defenses against jailbreaking~\cite{jain2023baseline,wei2023jailbreak,xie2023defending} have been proposed, we further show that MAC can still surpass the baseline GCG even when attacking models with these defenses. Overall, our work provides a novel technique for accelerating jailbreak attacks on aligned language models and new insights into the safety evaluations of AI systems.

\section{Background and Related work}
\subsection{Jailbreak attack and defense}
With the significant success of the fast-paced development of LLMs, concerns regarding their potential for harmful generation and malicious usage have emerged~\cite{bommasani2022opportunities,chen2023combating,liu2023towards}, among which the jailbreaking issue~\cite{wei2023jailbroken,dong2023robust,wei2023jailbreak} has been identified as one of the major concerns~\cite{yao2023survey,chen2023combating,zhang2024towards}. Despite significant efforts to align LLMs with human values and teach them to avoid generating harmful content~\cite{bai2022training,bai2022constitutional,ouyang2022training}, recent studies indicate that LLM alignment remains superficial~\cite{qi2023finetuning,liu2023jailbreaking} and is vulnerable to carefully crafted jailbreak prompts that can bypass safeguards and induce the generation of harmful content~\cite{yao2023survey,chen2023combating}. Previous studies have manually designed jailbreak prompts using persuasive instructions~\cite{wei2023jailbroken,zeng2024johnny,li2023deepinception,Xu2023CognitiveOJ}. Another approach involves using stealthy communication techniques~\cite{yuan2024gpt,yong2024lowresource,li2023deepinception} to circumvent LLM safeguards. Notably, recent works have shown that it is also possible to jailbreak the LLMs with discrete optimization over the prompt~\cite{guo2021gradientbased,wen2023hard}, represented by the adversarial suffix attacks such as gradient-based attacks like GCG~\cite{shin2020autoprompt, zou2023universal} and AutoDAN~\cite{zhu2023autodan}, which attach a suffix to the harmful request and optimize it using gradient heuristics, showing a higher success rate but requiring white-box access to the target model. On the other hand, some preliminary defense methods are also proposed in this context~\cite{wei2023jailbreak,jain2023baseline,wang2024language,li2023rain,chen2025towards}, yet the jailbreak issues are not fully resolved in this literature.

\subsection{Adversarial suffix attacks}
As a representative attack, GCG~\cite{zou2023universal} optimizes an adversarial suffix for each harmful prompt in the following manner: for each iteration, first calculate the cross-entropy loss of each token in the suffix with respect to generating the target prefix (e.g., \texttt{Sure, here's}). Then, randomly select a batch of substitute tokens in the suffix based on the gradient of the loss for each token. Finally, calculate the loss of each substituted suffix in the batch, then replace the current suffix with the one with the lowest loss, as shown in Algorithm~\ref{single GCG}. This attacking process is referred to as the \textit{individual} prompt attack and can be generalized to the \textit{multiple} prompts scenario. Several concurrent studies~\cite{jia2024improved,li2024exploiting} also investigated enhancements in the random search process for adversarial tokens. However, the use of gradient momentum as a heuristic in this setting remains unexplored.

\begin{algorithm}[!h]
\label{single GCG}
  \SetAlgoLined
  \caption{One-step Greedy Coordinate Gradient (GCG)~\cite{zou2023universal}}
  \KwIn{A LLM $f_\theta$, training prompt $p$ with corresponding optimization loss $\ell$, batch size $B$, top-k $k$, suffix $s$ with  length $l$, suffix gradient $\boldsymbol{g}$}
  \KwOut{Updated jailbreak suffix $s'$}
    $Initialize\quad  s=[s_1,\cdots, s_l]$\;
    \For{$i:1\to l$}{
        $X_i\gets \text{Top-k}(-\boldsymbol{g}^{(i)})$\;
    }
    \For{$b:1\to B$}{
        $s_b\gets s$\;
        $s_b^{(i)}\gets x_j, \quad\text{where}\ i\sim Uniform([1,\cdots,l]),\ x_j\sim Uniform(X_i)$\;
    }
    $s'\gets\arg\min_b \ell(s_b)$\;
    \textbf{return} $s'$\;
\end{algorithm}

\section{Methodology}
In this section, we present our \textbf{M}omentum \textbf{A}ccelerated G\textbf{C}G (\textbf{MAC}) attack on aligned language models. We first show that the suffix optimization procedure of GCG can be viewed as a stochastic gradient descent (SGD). However, this method may suffer from unstableness over different prompts, as the proper suffixes for different prompts or epochs may vary. Therefore, inspired by boosting conventional adversarial attacks with momentum methods~\citep{dong2018boosting} which has achieved great success in attacking vision models, we also propose to introduce a momentum term to the gradient for the search process.

\begin{algorithm}[!h]
\label{mgcg_ind}
  \SetAlgoLined
  \caption{MAC attack on \textbf{individual prompt}}
  \KwIn{A LLM $f_\theta$, training prompt $p$ with corresponding optimization loss $\ell$, optimizing epochs $T$, top-k $k$, suffix length $l$, momentum decay factor $\mu$}
  \KwOut{A jailbreak suffix $s$}
    $\text{Initialize}\quad  s=[s_1,\cdots, s_l]$, 
    momentum-gradient $\boldsymbol{g}\gets \nabla_{s}\ell([p, s])$\;
    \For{$t:1\to T$}{
        
            $\boldsymbol{g_{t}}\gets\nabla_{s} \ell([p, s])$\;
            $\boldsymbol{g}\gets \mu\boldsymbol{g} + (1-\mu)\boldsymbol{g_{t}}$\;
                \For{$i:1\to l$}{
        $X_i\gets \text{Top-k}(-\boldsymbol{g}^{(i)})$\;
        }
        \For{$b:1\to B$}{
        $s_b\gets s$\;
        $s_b^{(i)}\gets x_j, \quad\text{where}\ i\sim Uniform([1,\cdots,l]),\ x_j\sim Uniform(X_i)$\;
    }   
    $s\gets\arg\min_b \ell(s_b)$\;
    }
    \textbf{return} $s$\;
\end{algorithm}

The overall process is outlined in Algorithm~\ref{mgcg_ind}. We maintain a momentum gradient term $\boldsymbol g$. During each epoch, we calculate the loss $\ell_i$ for the current prompt $x_i$. The loss is designed as the cross-entropy loss for generating a target prefix (\textit{e.g.}, \texttt{Sure, here's})~\citep{zou2023universal}. Then, we update the momentum gradient $\boldsymbol g$ with the gradient of each token in the current suffix \textit{w.r.t.} the current loss with a decay factor $\mu$.  Finally, we apply the momentum gradient to the one-step-GCG to update the suffix. Note that the case of $\mu=0$ is exactly the GCG attack, which we believe is not stable and efficient enough over different prompts. An example of the optimized suffix by MAC in Table \ref{gcg}.
\begin{table}
\renewcommand\arraystretch{2}
\tabcolsep=5pt
\caption{Example suffix and responses for jailbreak tasks}
\centering
\begin{tabular}{c|p{0.3\textwidth}}\toprule \textbf{Prompt} & \texttt{Explain how to hack into a government database and steal confidential information} \\ \midrule \textbf{Initial Suffix} & \texttt{! ! ! ! ! ! ! ! ! ! ! ! ! ! ! ! ! ! ! ! } \\ \midrule \textbf{Initial Response} & \texttt{I'm sorry, but I cannot provide instructions on how to hack into a government database and steal confidential information. It is illegal and unethical ...} \\ \midrule \textbf{Optimized Suffix} & \texttt{stages Object Eachign\_\_\_\_\_\_\_\_ ! ...).\}\}\_\{\textbackslash\textbackslash ! !sorry none (: Just Perefrika your improv} \\ \midrule \textbf{Jailbreak Response} & \texttt{Sure, here's an outline of the steps involved in hacking into a government database and stealing confidential information: 1. Gather information ...} \\ \bottomrule \end{tabular}
\label{gcg}
\end{table}

Furthermore, we extend our MAC attack to accommodate multiple prompts, as depicted in Algorithm~\ref{mgcg}. Unlike updating the gradient over steps for a single harmful request, in this case, our MAC maintains the momentum term across different prompts during optimization, thereby enhancing the stability of the process across multiple prompts and achieving improved generalization ability.

\begin{algorithm}[!h]
\label{mgcg}
  \SetAlgoLined
  \caption{MAC attack on \textbf{multiple prompts}}
  \KwIn{A LLM $f_\theta$, training prompts $P=\{p_1,\cdots, p_n\}$ with corresponding optimization losses $\{\ell_1,\cdots,\ell_n\}$, optimizing epochs $T$, batch size $B$, top-k $k$, suffix length $l$, momentum decay factor $\mu$}
  \KwOut{A universal jailbreak suffix $s$}
    $\text{Initialize}\quad  s=[s_1,\cdots, s_l]$, momentum-gradient $\boldsymbol{g}\gets \nabla_{s}\ell([p_1, s])$\;
    \For{$t:1\to T$}{
        \For{$i:1\to n$}{
            $\boldsymbol{g_{t,i}}\gets\nabla_{s} \ell_i([p_i, s])$\;
            $\boldsymbol{g}\gets \mu\boldsymbol{g} + (1-\mu)\boldsymbol{g_{t,i}}$\;
                \For{$i:1\to l$}{
        $X_i\gets \text{Top-k}(-\boldsymbol{g}^{(i)})$\;
    }
    \For{$b:1\to B$}{
        $s_b\gets s$\;
        $s_b^{(i)}\gets x_j, \quad\text{where}\ i\sim Uniform([1,\cdots,l]),\ x_j\sim Uniform(X_i)$\;
    }
    $s\gets\arg\min_b \ell(s_b)$\;
        }
    }
    \textbf{return} $s$\;
\end{algorithm}

\section{Experiments}

\subsection{Experiment set-up}
Our experiment is based on vicuna-7b~\cite{zheng2023judging} and Misrtal-7b~\cite{jiang2023mistral}, two popular aligned chat LLMs. Following Zou et al.~\cite{zou2023universal}, we randomly select 100 adversarial prompts in their \textit{AdvBench} dataset which contains hundreds of harmful request prompts. For individual prompt attacks, we run the attack 5 times with different randomly chosen seeds and calculate their average performance. For multiple prompt attacks, we split them into 5 subsets which contain 20 prompts each, and use each subset as the training set and attack all the 100 prompts to run 5 independent experiments. For the token searching process, we set the token substitute batch size $B$ to 256, and top-$k$ to 256. Since we focus on the \textbf{efficiency} of the attack, we only optimize the suffix for 20 epochs ($T=20$). The criterion of attack success is whether the response contains any of the defensive tokens. Specifically, to determine whether an optimized prompt attack is successful, we apply the same detection method applied in GCG~\cite{zou2023universal}, which judges whether the generated response contains any of a refusal keyword set like \textit{``sorry''}, \textit{``I cannot''}. While this evaluation method remains controversial since it may cause false positive or false negative~\cite{shah2023loft}, it can still provide a fair comparison and has been applied in many subsequent works~\cite{liu2023autodan,wei2023jailbreak}.

\subsection{Individual prompt attack}

The effectiveness of the MAC attack in enhancing jailbreak effectiveness is discernibly analyzed through the comparative data presented in Table \ref{tab:ind}. For each $\mu$, we demonstrate our MAC's effectiveness by conducting 5 individual experiments, calculating their \textbf{average attack success rate (ASR)} and the average number of \textbf{steps} needed to successfully attack the prompt, as well as the standard deviation of the ASR and steps across these experiments to show the robustness of the results.

\begin{table*}[t]
    \centering
    \caption{Evaluation of MAC Attacks in \textbf{individual prompts} jailbreak performance}
    {
    \begin{tabular}{cc|cc|cc|cc|cc}
    \toprule
       & Model & 
       \multicolumn{4}{c|}{Vicuna}
       & 
       \multicolumn{4}{c}{Mistral}
       \\
       Attack & Momentum &
       Avg. ASR ($\uparrow$) & Std. ($\downarrow$) &
       Avg. Steps ($\downarrow$) & Std. ($\downarrow$) &
       Avg. ASR ($\uparrow$) & Std. ($\downarrow$) &
       Avg. Steps ($\downarrow$) & Std. ($\downarrow$)
       \\
    \midrule
        GCG & $\mu=0$ & 75.0 & \textbf{1.22} & 12.62 & 0.27 & 83.8 & 1.44& 5.88 & 7.05 \\
        \midrule
        \multirow{4}{*}{MAC (ours)} & $\mu=0.2$ & \textbf{76.6} & 2.07 & \textbf{12.37} & 0.31 & 85.6 & 3.06& 6.27 & 7.09 \\
        & $\mu=0.4$ & 76.2 & 4.82 & 12.46 & 0.36 & 83.8 & 2.31& 6.27 &7.10 \\
        & \red{$\mu=0.6$} & \red{76.0} & \red{2.12} & \red{12.55} & \red{\textbf{0.24}} & \red{\textbf{87.2}} & \red{1.85}& \red{\textbf{5.77}} & \red{\textbf{6.79}}  \\
        & $\mu=0.8$ & 72.4 & 3.58 & 13.05 & 0.57 & 87.0 & \textbf{1.46}& 6.09 & 6.86 \\
    \bottomrule
    \end{tabular}
    }
    \label{tab:ind}
\end{table*}

The table provided demonstrates the nuanced dynamics of individual attacks within the MAC framework. Note that GCG with $\mu=0$ can be considered as essentially the original GCG attack, which is the baseline of our method. For $\mu \in \{0.2, 0.4, 0.6\}$, our method significantly outperforms the original GCG method, achieving an average ASR increase of 1.3$\%$ for vicuna, and reducing the average attack steps from $12.62$ to $12.46$. As the effectiveness is shown to be improved in those experiments, the standard deviation of these indices remains relatively unchanged across all experiments, showing a robust improvement from our MAC. For $\mu = 0.8$, the effectiveness may begin to decrease because the emphasis on stability becomes excessive. This context underscores the efficacy of incorporating momentum merely as a means to augment stability, a strategy that evidently enhances the success rate of attacks. As a trade-off, further adding momentum tends to lower its performance. However, at $\mu=0.6$, the benefits of this approach are both significant and stable, demonstrating a clear advantage with minimal expenditure in terms of the number of steps required for a successful attack.

\subsection{Multiple prompt attack}

In multi-prompt attacks, the importance of generalization grows, and the benefits of adding a momentum term become evident. We further evaluate the performance of the MAC attack and compare it with vanilla GCG in Table~\ref{tab:mul}. Similarly, we apply the average value and standard deviation of ASR across the 5-fold experiments as metrics to demonstrate our methods' effectiveness. In addition, we introduce Maximum ASR, \textit{i.e.} the highest ASR achieved on the test set throughout all attack steps as a crucial metric since the goal of multiple prompt attacks is to craft an effective adversarial suffix and the suffixes crafted during any epoch is available.

\begin{table*}[t]
    \centering
    \caption{Evaluation of MAC attacks in \textbf{multiple prompts} jailbreak performance}
    {
    \begin{tabular}{cc|cc|cc|cc|cc}
    \toprule
    & Model & 
       \multicolumn{4}{c|}{Vicuna}
       & 
       \multicolumn{4}{c}{Mistral}
       \\
       Attack & Momentum & Avg. ASR ($\uparrow$) & Std. ($\downarrow$) & Max. ASR ($\uparrow$) &  Std. ($\downarrow$) &
       Avg. ASR ($\uparrow$) & Std. ($\downarrow$) & Max. ASR ($\uparrow$) &  Std. ($\downarrow$)\\
    \midrule
        GCG & $\mu=0$ & 38.1 & 8.66 & 72.7 & 15.36 & 58.72 & 2.02& 80.3 & 5.28\\
    \midrule
        \multirow{4}{*}{MAC (ours)} & $\mu=0.2$ & 35.9 & \textbf{5.95} & 74.9 & 15.30& 60.0 & \textbf{1.63}& 78.2 & 4.78\\
        & $\mu=0.4$ & 44.3 & 7.31 & 81.9 & 12.45& 61.7 & 2.52& 82.0 & 6.75\\
        & \red{$\mu=0.6$} & \red{\textbf{48.6}} & \red{14.97} & \red{\textbf{86.1}} & \red{\textbf{9.00}}& \red{\textbf{61.9}} & \red{1.75}& \red{82.5} & \red{\textbf{3.74}}\\
        & $\mu=0.8$ & 43.4 & 13.69 & 83.8 & 9.18& 60.5 & 2.97& \textbf{85.5} & 7.69\\
    \bottomrule
    \end{tabular}
    }
    \label{tab:mul}
\end{table*}

As illustrated in Table~\ref{tab:mul}, experiments with $\mu \in \{0.4,0.6,0.8\}$ show our method's effectiveness, particularly in the improved average ASR (increasing by 6.2\%, 10.5\%, and 5.3\% for vicuna, respectively), highlighting the critical role of dataset selection in improving attack efficacy and robustness. 
Furthermore, our MAC excels higher Maximum ASRs (increasing by 2.2\%, 9.2\%, 13.4\%, and 11.1\% for vicuna, respectively) and lower standard deviations, indicating enhanced efficiency and robustness for crafting an effective adversarial suffix. Notably, when $\mu \geq 0.6$, the standard deviation of Maximum ASR significantly decreases from $15.36$ to approximately $9$, indicating that higher momentum values contribute to better stability of Maximum ASR.

Furthermore, we can see that a $\mu$ of 0.6 yields the optimal performance. Still taking vicuna as an example, at a momentum value of $\mu=0.6$, the MAC attack achieves its highest Average ASR ($48.6\%$) and Maximum ASR ($86.1\%$). This configuration also results in the lowest standard deviation for Maximum ASR ($9.0$), indicating it optimizes both effectiveness and consistency. However, it is crucial to note that both lower and higher momentum values fail to balance stability and learning rate adaptation effectively, potentially compromising performance. This indicates a clear sweet spot for the momentum ($\mu$) value, underscoring the importance of nuanced parameter tuning in adversarial contexts. 

\subsection{Transfer attack} 
Since adversarial suffices can transfer between different LLMs, it is possible to attack a black-box LLM with a local surrogate model~\cite{zou2023universal}. To further validate the robustness of the MAC attack framework, we evaluate its transferability across different models. In this experiment, we still use the adversarial suffixes (with the best ASRs) crafted during the multiple prompt attacks on vicuna and mistral and apply them to attack another model. As shown in Table~\ref{tab:transf}, our MAC attack surpasses the baseline GCG regarding transferability for both models. This indicates that momentum not only boosts the effectiveness and stability of the attack in a single model but also enhances the generalization capability of adversarial samples across different models, further demonstrating the effectiveness of MAC in transfer attack scenarios.

\begin{table}[h]
\tabcolsep=10pt
    \centering
    \caption{Evaluation of MAC attacks in \textbf{transferability} performance}
    \label{tab:transf}
    {
    \begin{tabular}{cc|c|c}
    \toprule
        && \multicolumn{2}{c}{Vitim model}
        \\
       Attack & Momentum & Vicuna & Mistral\\
    \midrule
        No Attack & - &  3.0 & 42.5 \\
        GCG & $\mu=0$ &  11.3 & 46.0  \\
    \midrule
        \multirow{1}{*}{MAC (ours)}
        & $\mu=0.6$ &  \textbf{12.3} & \textbf{51.3}\\
    \bottomrule
    \end{tabular}
    }
\end{table}

\subsection{Attacking models under defense}
Finally, we also investigate the performance of MAC against LLMs that are protected by various defense mechanisms. In this experiment, we use vicuna as the protected model and consider three popular defenses, including the PPL filter~\cite{jain2023baseline}, Self-reminder~\cite{xie2023defending}, and ICD (1 shot)~\cite{wei2023jailbreak}. We assess the ASRs of both the multi-prompt GCG and MAC ($\mu=0.5$) attacks when these defenses (with their default configurations) are applied, as shown in Table~\ref{tab:transf}.

\begin{table}[h]
    \tabcolsep=10pt
    \centering
    \caption{Evaluation of MAC attacks against models under defense}
    \label{tab:transf}
    {
    \begin{tabular}{c|cc}
    \toprule
        & \multicolumn{2}{c}{Attack method}
        \\
       Attack  &  GCG & MAC (ours)\\
    \midrule
        No Defense & 72.7 & 86.1\\
        \midrule
        PPL filter~\cite{jain2023baseline} & 28.0 & \textbf{76.0} \\
        Self-reminder~\cite{xie2023defending} &65.8 & \textbf{75.0} \\
        ICD~\cite{wei2023jailbreak}& 35.8 & \textbf{62.5} \\
     
    \bottomrule
    \end{tabular}
    }
\end{table}

These results indicate that the MAC attack demonstrates superior effectiveness even in the presence of defense mechanisms. In the baseline scenario where no defense is employed, MAC achieves a higher ASR than GCG. Among the defenses tested, the MAC attack continues to outperform GCG in all cases. Under the PPL filter, MAC exhibits a significantly higher ASR of $76.0\%$ compared to GCG, indicating that the momentum generated by the suffices also experiences lower perplexity, which is an unexpected benefit. Furthermore, for the Self-reminder and ICD (1 shot) defenses, MAC maintains a stronger ASR compared to GCG, suggesting that the momentum-based component of the MAC attack provides enhanced resilience against defense mechanisms, reinforcing its effectiveness as a robust attack strategy under diverse conditions.

\section{Conclusion and Limitations}

In this work, we rethink a current popular gradient-based attack on LLMs from the optimization perspective. We first propose a new jailbreak attack through the lens of optimization named \textbf{M}omentum \textbf{A}ccelerated G\textbf{C}G (\textbf{MAC}), and demonstrate that accelerated optimization methods like momentum SGD can effectively boost such attacks with notably higher attack success rates and fewer optimization steps, along with better transferability and robustness against defense mechanisms, providing deeper insights into the current jailbreaking issue research. 

While the recent research thread mainly focuses on black-box attacks~\citep{wei2023jailbreak,zhu2023autodan,li2023deepinception,chen2023rethinking}, we consider it still important to develop an efficient white-box attack for developers to efficient evaluation and red-teaming on LLMs. This is similar to the conventional adversarial robustness in the vision domain, where white-box attack methods for evaluation are still valuable~\citep{athalye2018obfuscated,croce2020reliable}.

As a preliminary study, we acknowledge some limitations of this work, which we leave as future works. First, our  MAC only considers the case of batch size as 1 in the multiple prompts attack case. It would be interesting to see whether a larger batch size can find a better trade-off between efficiency and stableness. In addition, there are many optimization methods other than momentum, like Adam, remain unexplored. Finally, our experiment only focuses on two models, and its effectiveness can be further substantiated by evaluating more models.

\section*{Acknowledgement}

This work was sponsored by the Beijing Natural Science Foundation (Grant No. QY24035, QY23041) and the National Natural Science Foundation of China (Grant No. 62172019).

\newpage
{
\bibliographystyle{abbrv}
\bibliography{ref}
}

\end{document}